\documentclass{article}
\usepackage{arxiv}
\usepackage[utf8]{inputenc}
\usepackage{amssymb,amsthm}
\usepackage{enumitem}
\usepackage{microtype}
\usepackage{idxlayout}
\usepackage{lmodern}
\usepackage{float}
\usepackage{url}
\usepackage{amsthm}
\usepackage{graphicx}
\usepackage{mathtools}
\usepackage{longtable,booktabs}
\usepackage{subcaption}
\usepackage{fullpage}
\usepackage{times}
\usepackage[rightcaption]{sidecap}
\sidecaptionvpos{figure}{t}
\usepackage{fancyhdr,graphicx,amsmath,amssymb}
\usepackage{wrapfig,lipsum,booktabs}
\usepackage[ruled,vlined, linesnumbered]{algorithm2e}
\include{pythonlisting}

\makeatletter
\newcommand{\printfnsymbol}[1]{%
  \textsuperscript{\@fnsymbol{#1}}%
}
\makeatother

\begin{document}

\title{Policy Optimization Reinforcement Learning with Entropy Regularization}

\author{
  Jingbin Liu \thanks{equal contribution} \\
  CreateAmind\\
  \And
 Xinyang Gu\printfnsymbol{1} \\
  CreateAmind\\
     \And
 Shuai Liu \\
  CreateAmind\\
}

\maketitle

\begin{abstract}
Entropy regularization is an important idea in reinforcement learning, with great success in recent algorithms like Soft Q Network (SQN) and Soft Actor-Critic (SAC1). In this work, we extend this idea into the on-policy realm. We propose the soft policy gradient theorem (SPGT) for on-policy maximum entropy reinforcement learning. With SPGT, a series of new policy optimization algorithms are derived, such as SPG, SA2C, SA3C, SDDPG, STRPO, SPPO, SIMPALA and so on. We find that SDDPG is equivalent to SAC1. For policy gradient, the policy network is often represented as a Gaussian distribution with a global action variance, which damages the representation capacity. We introduce a local action variance for policy network and find it can work collaboratively with the idea of entropy regularization. Our method outperforms prior works on a range of benchmark tasks. Furthermore, our method can be easily extended to large scale experiment with great stability and parallelism.
\end{abstract}


\section{Introduction}\label{header-n1}
Deep reinforcement learning is interesting and exciting because good reinforcement learning algorithms can learn the solutions for us in an automatic way. The combination of reinforcement learning and high-capacity function approximators such as neural networks holds the promise of solving a large range of decision making and control tasks. Great success has been achieved in traditional board games like Go and Chess to complex strategic games like DOTA2 and StarCraftII \cite{AlphaZero}\cite{Dota2}\cite{AlphaStar}. Despite the remarkable achievements, most of these methods lose effectiveness in real-world applications, hindered by all kinds of challenges\cite{world_model}.

First of all, most of current methods can suffer from the sub-optimum problem, usually with a solution doing well in the short term, but under-performing in the long run. Secondly, on-policy reinforcement learning methods tend to have low sample-efficiency. Take the standard baseline environments Atari57 for example, billions of interaction steps are needed to solve each one of the environments\cite{R2D2}\cite{APEX}\cite{IMPALA}. Many researches tend to obtain state-of-the-art results with more and more computing resources, not concerned with sample-efficiency. Last but not least, current reinforcement learning algorithms are often extremely brittle with respect to their hyper-parameters. Fore sparse reward environment, reward design can also be a problem\cite{design}. Parameter tuning and reward design can be time-consuming and harm reproducibility.

In this paper, we argue that in the reinforcement learning domain, explore-exploit trade-off, or explore-exploit paradox, is the key problem that needs more attention if we want to achieve stable training with high efficiency. All kinds of mechanisms that are able to interpret the connection between exploration and exploitation can be helpful. Entropy regularization is an important idea resorting to explore-exploit balance in reinforcement learning \cite{SQL}\cite{SAC}\cite{SAC1}. In this work, we extend this idea into the on-policy reinforcement learning domain and improve the efficiency of the policy-gradient based algorithms. We propose the soft policy gradient theorem and construct soft policy gradient algorithms according to the theorem. Corresponding to common policy gradient based algorithms like PG, A2C, A3C, DDPG, PPO, IMPALA etc., we develop their soft counterparts. In traditional on-policy algorithms, a small entropy bonus reward is often added to encourage exploration. The entropy bonus should be small and tuned for stable training. With the SPGT, we show that the entropy bonus used in on-policy algorithms is a natural consequence of the theorem’s self-consistency and can be derived in a rigorous manner in the entropy-regularized reinforcement learning framework.

\section{Background}\label{header-n2}

\subsection{MDP}\label{header-n21}

The key problem of RL is searching for a policy that maximizes the accumulate future rewards 
in a Markov Decision Process \((\text{MDP})\) defined by
the tuple \((\mathcal{S,A,P,R})\) \cite{planet}. 

\begin{itemize}
\item
  \(\mathcal {S}\) represents a set of states
\item
  \(\mathcal {A}\) represents a set of actions, 
\item
  \(\mathcal {P:S\times A\to P(S)}\) stands for the transition function which maps
  state-actions to probability distributions over next states
  \(\text{P}(s^{\prime}|s,a)\)
\item
  \(\mathcal {R}:\mathcal{S\times A\times S}\to \mathbb{R} \) corresponds to the
  reward function, with \(r_t=R(s_t,a_t,s_{t+1})\)
\end{itemize}

Within this framework, the agent acts in the environment according to a policy $a  \sim \pi(\cdot|s)$.
the environment changes to a new state following $s^{\prime} \sim \mathcal{P}(\cdot|s, a)$ and provides a reward $r$. The agent iterates its policy based on the information gathered via interacting with the environment to obtain an optimal policy:
\begin{equation}
    \pi^* = \arg \underset{\pi}\max \underset{\tau \sim \pi} {\text{E}} \bigg[{ \sum_{t=0}^{\infty} \gamma^t  r(s_t, a_t, s_{t+1}) }\bigg]
\end{equation}{}

\subsection{Policy Gradient}

Policy gradient methods maximize the expected return based on the policy gradient theorem, by directly computing an estimate of the gradient of policy parameters. With the help of deep neural networks, the policy gradient algorithms have become an important model-free reinforcement learning. Such methods are also attractive because they don’t require an explicit model of the environment.

The gradient of the objective $J(\pi_\theta)$ has the form $\text{E}_{\tau \sim \pi_\theta}\Big[\sum_{t=0}^T \triangledown_\theta
\log \pi_\theta (a_t|s_t) A^{\pi_\theta}(s_t,a_t) \Big]$, where $A^{\pi_\theta}(s_t,a_t)$ is the advantage estimate.
For the simplest case, the REINFORCE algorithm takes the sample return of a trajectory $A^{\pi}=\sum_{t=0}^T r(s_t,a_t)$.
However, it suffers from high variance in the gradient estimates. Instead value function baselines are used in modern policy algorithms, i.e. Actor-Critic algorithms: $A^{\pi}=\sum_{t=0}^T r(s_t,a_t)-V_\theta(s_t)$. Furthermore, Generalized advantage estimation are introduced to balance the estimate variance and bias. 

Policy gradient algorithms are typically on-policy algorithms, distributed policy gradient algorithms can suffer severe policy lag because of asynchronous sampling and training. Two major classes of techniques are designed to cope with off-policy learning. The first approach is to apply trust region methods by staying close to the behavior policy during training, such as Trust region policy optimization (TRPO) \cite{TRPO} and Proximal policy optimization (PPO) \cite{PPO}. Another idea is to use importance sampling to correct the targets for the value function to improve the approximation of the discounted sum of rewards under the target policy, such as V-trace algorithm introduced in Importance-weighted actor-learner architectures (IMPALA) \cite{IMPALA}.

\subsection{Maximum Entropy Reinforcement Learning}\label{header-n22}

The maximum entropy reinforcement learning generalizes the standard objective with an entropy term, such that the optimal policy additionally aims to maximize its entropy at each visited state \cite{SQL}. The temperature parameter $\alpha$ that determines the relative importance of the entropy term versus the reward is account for the explore-exploit balance.
\begin{equation}
    \pi^* = \arg \underset{\pi}\max \underset{\tau \sim \pi} {\text{E}} \bigg[{ \sum_{t=0}^{\infty} \gamma^t \bigg( r(s_t, a_t, s_{t+1}) + \alpha H\left(\pi(\cdot|s_t)\right) \bigg)}\bigg].
\end{equation}{}

The temperature parameter $\alpha$ can be tuned to encourage exploration. When $\alpha$ is large, the policy will be more stochastic, on the contrary, when $\alpha$ is small, the policy will become more deterministic. In the limit $\alpha \rightarrow 0$, We can recover exactly the standard reinforcement learning. The maximum entropy objective has a number of conceptual and practical advantages. The policy is incentivized to explore more widely, capturing multiple modes of near-optimal behavior while giving up on unpromising avenues. Unlike the standard reinforcement learning, in problem settings where multiple actions seem equally attractive, the policy will commit equal probability mass to those actions instead of collapsing into one action randomly.

In the entropy-regularized framework, \(v_{\pi}\) and \(q_{\pi}\) should be modified to include the entropy term: 
\begin{align}
    v_{\pi}(s) &= \underset{\tau \sim \pi} {\text{E}} \bigg[{ \left. \sum_{t=0}^{\infty} \gamma^t \bigg( r(s_t, a_t, s_{t+1}) + \alpha H\left(\pi(\cdot|s_t)\right) \bigg) \right| s_0 = s}\bigg]  \label{eq:v} \\
    q_{\pi}(s,a) &= \underset{\tau \sim \pi} {\text{E}}\bigg[{ \left. \sum_{t=0}^{\infty} \gamma^t \bigg( r(s_t, a_t, s_{t+1}) + \alpha  H\left(\pi(\cdot|s_t)\right) \bigg) \right| s_0 = s, a_0 = a}\bigg]
    \label{eq:q}
\end{align}
With equation(\ref{eq:v}, \ref{eq:q}), we can draw the connection between \(v_{\pi}\) and \(q_{\pi}\). Meanwhile, we have the Bellman equation for \(v_{\pi}\) and \(q_{\pi}\):
\begin{align}
    v_{\pi}(s) &= \underset{a \sim \pi}{\text{E}}[{q_{\pi}(s,a)}] \\
    q_{\pi}(s,a) &= \underset{s' \sim P}{\text E}[{r(s,a,s') + \alpha  H\left(\pi(\cdot|s)\right) + \gamma v_{\pi}(s')}]
\end{align}

\section{The Soft Policy Gradient Theorem}\label{header-n3}

To keep the notation simple, we leave it implicit in all cases that $\pi$ is a function of $\theta$,
and all gradients are also implicitly concerning $\theta$. The discount factor $\gamma$ is omitted.
First note that the gradient of the state-value function can be written in terms of the action-value function as
\begin{align}
    \bigtriangledown v_{\pi}(s) 
    &= \bigtriangledown \left[ \sum_a \pi(a|s)q_{\pi}(s,a)\right] \\
    &= \sum_a \left[ q_{\pi}(s,a) \bigtriangledown \pi(a|s) + \pi(a|s) \bigtriangledown q_{\pi}(s,a) \right] \\
    &= \sum_a \left[ q_{\pi}(s,a) \bigtriangledown \pi(a|s) + \pi(a|s) \bigtriangledown   \sum_{s^\prime,r} p(s^\prime,r|s,a)(r_\pi(s,a)+v_\pi(s^\prime)) \right]\\
    & (\text{where} \: r_\pi(s,a) = r + \alpha H(\pi(\cdot|s))) \nonumber \\
    & = \sum_a \left[ q_{\pi}(s,a) \bigtriangledown \pi(a|s) + \pi(a|s) \sum_{s^\prime} p(s^\prime|s,a) \bigtriangledown  (\alpha H(\pi(\cdot|s)) +v_\pi(s^\prime)) \right] \\
    &= \sum_a \left[ q_{\pi}(s,a) \bigtriangledown \pi(a|s) + \pi(a|s) \alpha\bigtriangledown  H(\pi(\cdot|s)) + \pi(a|s) \sum_{s^\prime} p(s^\prime|s,a) \bigtriangledown  v_\pi(s^\prime) \right]   \nonumber\\
    &=\sum_a \left[ q_{\pi}(s,a) \bigtriangledown \pi(a|s) +\alpha\bigtriangledown  H(\pi(\cdot|s)) + \pi(a|s) \sum_{s^\prime} p(s^\prime|s,a) \bigtriangledown  v_\pi(s^\prime) \right]\\
    &\cdot\cdot\cdot \left(\bigtriangledown v_\pi(s^\prime) = \sum_{a^\prime} \left[ q_{\pi}(s^\prime,a^\prime) \bigtriangledown \pi(a^\prime|s^\prime) +\alpha\bigtriangledown  H(\pi(\cdot|s^\prime)) + \pi(a^\prime|s^\prime) \sum_{s^{\prime\prime}} p(s^{\prime\prime}|s^\prime,a^\prime) \bigtriangledown  v_\pi(s^{\prime\prime}) \right]\right) \nonumber \\
    &  \cdot\cdot\cdot  (\text{unrolling}) \nonumber \\
    &=\sum_{x \in \mathcal{S}} \sum_{k=0}^{\infty} \Pr (s \rightarrow x,k,\pi)\sum_a[q_{\pi}
    (x,a)\triangledown \pi(a|x) + \alpha \triangledown H(\pi(\cdot|x))],
    \label{eq:value}
\end{align}

where $\Pr (s \rightarrow x,k,\pi)$ is the probability of transitioning from state $s$ to state $x$ in $k$ steps under policy $\pi$. It is then immediate that
\begin{align}
    \triangledown J(\theta)
    &=\triangledown v_{\pi}(s_0)\\
    &= \sum_s(\sum_{k=0}^{\infty}\Pr \Big(s_0 \rightarrow s,k,\pi)\Big)\sum_a\Big[q_{\pi}(s,a)\ \triangledown \pi(a|s) + \alpha \triangledown H(\pi(\cdot|s)) \Big]\\
    &= \sum_s \eta(s) \sum_a\Big[q_{\pi}(s,a)\ \triangledown \pi(a|s) + \alpha \triangledown H(\pi(\cdot|s)) \Big]\\
    &= \sum_s \eta(s') \sum_s \frac{\eta(s)}{\sum_{s'}\eta(s')} \sum_a \Big[q_{\pi}(s,a)\ \triangledown \pi(a|s) + \alpha \triangledown H(\pi(\cdot|s)) \Big]\\
    &\propto \sum_s \mu(s) \sum_a \Big[q_{\pi}(s,a) \triangledown \pi(a|s) + \alpha \triangledown H(\pi(\cdot|s)) \Big] \\
    &\text{where} \; \mu(s) = \sum_s \frac{\eta(s)}{\sum_{s'}\eta(s')} \nonumber\\
    &= \sum_s \mu(s) \sum_a \Big[q_{\pi}(s,a) - \alpha \log \pi(a|s) \Big] \triangledown \pi (a|s)
\end{align}

Finally, we have the soft policy gradient theorem as below:

\begin{align}
    \triangledown J(\theta) & \propto \sum_s \mu(s) \sum_a \Big[q_{\pi}(s,a) \triangledown \pi(a|s) + \alpha \triangledown H(\pi(\cdot|s)) \Big]  \label{scheme2} \\
    &\propto \sum_s \mu(s) \sum_a \Big[q_{\pi}(s,a) - \alpha \log \pi(a|s) \Big] \triangledown \pi (a|s)  \label{scheme1}
\end{align}

It seems crucial for the derivation that the entropy is with $s$ instead of $s^\prime$, i.e. $H(\pi(\cdot|s))$. If we use the definition  $H(\pi(\cdot|s^\prime))$ for the entropy regularization, $H(\pi(\cdot|s^\prime))$  should be the expectation of $ \pi(a|s) \sum_{s^\prime} p(s^\prime|s,a)$, but the formula don't change. 

It should be noted that within the policy gradient theorem, the entropy regularization term undermines the self-consistency of the theorem. While within the soft policy gradient theorem, the entropy regularization term is no longer a dispensable reward bonus for exploration, but a rigorous consequence of the soft policy gradient theorem's self-consistency. 
From Eq.(\ref{scheme2}), we can see that the standard policy gradient with entropy bonus and the soft policy gradient reinforcement learning have the same loss function. The only difference is that SPG has the entropy reward in the value function backup while PG does not, which can be seen in Eq.(\ref{eq:v}).

\subsection{SDDPG = SAC1}

There is a special case in the policy gradient algorithms. It is named Deep Deterministic Policy Gradient (DDPG) \cite{DDPG}. Here we show the connection between soft DDPG and SAC1 \cite{SAC1}. To achieve that, two major techniques are performed on policy gradient.

1. Reparametrization trick: (continuous action with Gaussian distribution as an example, also applicable to discrete case and other sample distributions)
\begin{align}
    &  \Tilde{\mu}_\theta(s) \sim \pi_\theta(\cdot|s) \nonumber \\
    &  \Tilde{\mu}_\theta(s) = \mu_\theta(s) + \sigma_\theta(s) \mathcal{N}(0,1)   \nonumber \\
    &  \Tilde{\mu}^\epsilon_\theta(s) = \mu_\theta(s) + \sigma_\theta(s) \epsilon, \ \ \epsilon \sim \mathcal{N}(0,1)   \nonumber 
\end{align}
To keep the notation simple, we leave it implicit in all cases that $\pi$ is a function of $\theta$,
and all gradients are also implicitly concerning $\theta$. 
\begin{align}
    \bigtriangledown v_{\pi}(s) 
    &= \bigtriangledown \left[ \sum_a \pi(a|s)q_{\pi}(s,a)\right] \\
    &= \sum_a \left[ q_{\pi}(s,a) \bigtriangledown \pi(a|s) + \pi(a|s) \bigtriangledown q_{\pi}(s,a) \right] \\
    &= \sum_a \left[ q_{\pi}(s,a) \bigtriangledown \pi(a|s) + \pi(a|s) \bigtriangledown   \sum_{s^\prime,r} p(s^\prime,r|s,a)(r(s,a)+v_\pi(s^\prime)) \right]\\
    &\cdot\cdot\cdot \left( \sum_a  q_{\pi}(s,a) \bigtriangledown \pi(a|s) = \bigtriangledown_{\theta_\pi} {\text{E}}_{ a \sim \pi(a|s)}[ q_{\pi}(s,a)]= \bigtriangledown_{\theta_\mu} \text{E}_{ \epsilon \sim \mathcal{N}(0,1)}[ q_{\pi}(s,\Tilde{\mu}^\epsilon_\theta(s))] \right) \label{eq:eq1}\\
    &= \text{E}_{ \epsilon \sim \mathcal{N}(0,1)} [\bigtriangledown_{a}q_{\pi}(s,a)|_{a=\Tilde{\mu}^\epsilon_\theta(s)} \bigtriangledown_{\theta}\Tilde{\mu}^\epsilon_\theta(s)] + \sum_a \left[ \pi(a|s) \sum_{s^\prime} p(s^\prime|s,a) \bigtriangledown  v_\pi(s^\prime) \right]   \nonumber\\
    &  \cdot\cdot\cdot  (\text{unrolling}) \nonumber \\
    &=\sum_{x \in \mathcal{S}} \sum_{k=0}^{\infty} \Pr (s \rightarrow x,k,\pi)
    \text{E}_{ \epsilon \sim \mathcal{N}(0,1)} [\bigtriangledown_{a}q_{\pi}(s,a)|_{a=\Tilde{\mu}^\epsilon_\theta(s)}\bigtriangledown_{\theta}\Tilde{\mu}^\epsilon_\theta(s)],
    \label{eq:rpg}
\end{align}

where $\Pr (s \rightarrow x,k,\pi)$ is the probability of transitioning from state $s$ to state $x$ in $k$ steps under policy $\pi$. We have the reparametried policy gradient (RPG) as below:
\begin{align}
    \triangledown J(\theta)
    &\propto \sum_s \rho(s) \text{E}_{ \epsilon \sim \mathcal{N}(0,1)} [\bigtriangledown_{a}q_{\pi}(s,a)|_{a=\Tilde{\mu}^\epsilon_\theta(s)}\bigtriangledown_{\theta}\Tilde{\mu}^\epsilon_\theta(s)]
\end{align}
It's easy to verify that DDPG is a special case of RPG when $  \sigma_\theta(s) = 0 $:
\begin{align}
    \triangledown J(\theta)
    &\propto \sum_s \rho(s) [\bigtriangledown_{a}q_{\pi}(s,a)|_{a={\mu}_\theta(s)}\bigtriangledown_{\theta}{\mu}_\theta(s)]
\end{align}

\begin{figure}
    \centering
    \includegraphics[width=0.79\textwidth]{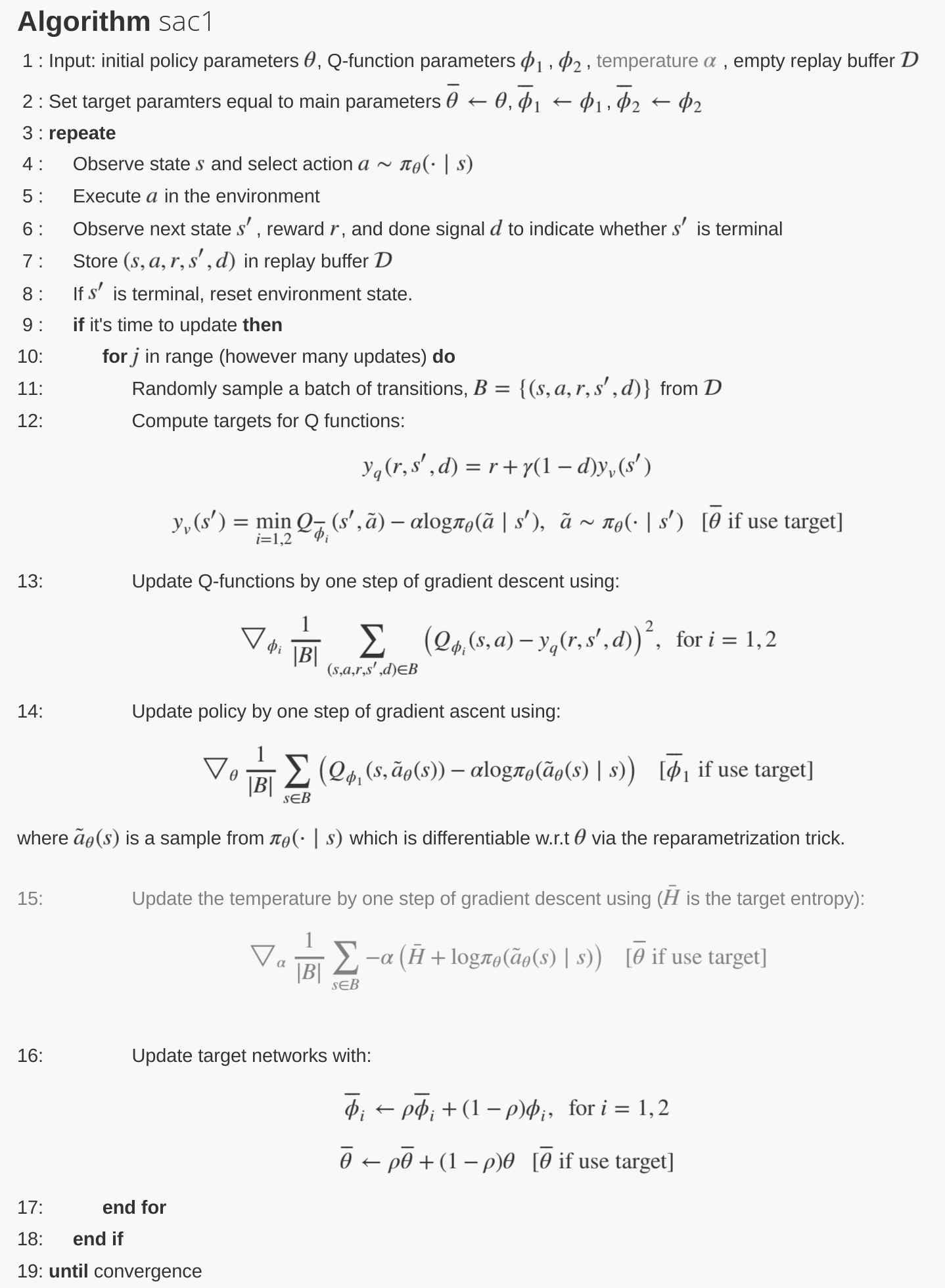}
    \caption{The SAC1 algorithm. There are two Q function networks and one policy network in the framework. Q function networks are updated with Q value backups and policy network is updated according to the learned Q function.}
    \label{fig:sac1}
\end{figure}

2. Entropy regularization: 

The above derivation are based on the standard reinforcement learning. Here we shift RPG to the maximum entropy reinforcement learning framework. Referring to the derivation of SPGT, we can easily implement the entropy regularization in RPG by proceeding with Eq.\ref{eq:eq1}:
\begin{align}
& \sum_a  q_{\pi}(s,a) \bigtriangledown \pi(a|s) + \alpha\bigtriangledown  H(\pi(\cdot|s)) \\
&=\bigtriangledown_{\theta_\pi}  \sum_a \pi(a|s) [q_{\pi}(s,a)-\alpha \text{log}  \pi(a|s)]  \\
& = \bigtriangledown_{\theta_\pi} {\text{E}}_{ a \sim \pi(a|s)}[ q_{\pi}(s,a)-\alpha \text{log}  \pi(a|s)]\\
&= \bigtriangledown_{\theta} \text{E}_{ \epsilon \sim \mathcal{N}(0,1)}[ q_{\pi}(s,\Tilde{\mu}^\epsilon_\theta(s))
-\alpha \text{log}  \pi_\theta(\Tilde{\mu}^\epsilon_\theta(s)|s)] \\
&= \bigtriangledown_{\theta}[ q_{\pi}(s,\Tilde{\mu}_\theta(s))
-\alpha \text{log}  \pi_\theta(\Tilde{\mu}_\theta(s)|s)], \ \ \Tilde{\mu}_\theta(s) \sim \pi_\theta(\cdot|s) \label{eq:sddpg}
\end{align}
Thus we have the SDDPG algorithm, which is exactly the SAC1 algorithm. The pseudo-code and implementation of the SAC1 algorithm are available at \url{https://github.com/createamind/DRL}. We can see that Eq.(\ref{eq:sddpg}) is exactly the policy update rule in Line 14 of Fig.\ref{fig:sac1}.

Although the derivations of the SAC1 algorithm and the SDDPG algorithm are completely different, the two algorithms are actually identical. The fact reveals the equivalence between soft Q-learning and soft policy gradient.

\subsection{The Soft Policy Gradient Algorithms}

With SPGT, the soft version of all the policy optimization algorithms in the standard reinforcement learning can be derived, such as SPG, SA2C, SA3C, SDDPG(SAC1), STRPO, SPPO, SIMPALA and so on.
As an example, we choose the popular on-policy algorithm PPO and modify it to the SPPO algorithm. 

In order to construct the SPPO algorithm, we should first define the policy gradient loss and value loss.  In the framework of entropy-regularized reinforcement learning, the reward have two components  $r =r^{\text{ext}} + r^{\text{int}}$ : the external reward $r^{\text{ext}}$ given by the environment  and the internal  $r^{\text{int}}$ defined according to the policy entropy. The objective for the value part is to minimize a square-error loss for value estimation:
\begin{equation}
    L^v_t(\phi) = (V_{\phi}(s_t) - V_t^\text{targ})^2 = (V_{\phi}(s_t) - \hat{R}_t)^2 \label{eq:Lv}
\end{equation}
The advantage estimator that look within $T$ timesteps can be written as:
\begin{equation}
    \hat{A}_t = \hat{R}_t - V(s_t) = r_t + \gamma r_{t+1} + \cdot\cdot\cdot +\gamma^{T-t-1} r_{T-1} + \gamma^{T-t} V(s_T)  - V(s_t) \label{eq:At}
\end{equation}
with
\begin{equation}
    r =r^{\text{ext}} + r^{\text{int}} = r^{\text{ext}} + \alpha H(\pi(\cdot|s)), \label{eq:At1}
\end{equation}
where $t$ specifies the time index in $[0, T]$, within a given T-timestpe truncated trajectory. Generalizred Advantage Estimation (GAE) is the generalized version of the normal advantage estimation. GAE can reduce to the above equation when $\lambda = 1$:
\begin{align}
    &\hat{A}_t  = \delta_t + (\gamma\lambda) \delta_{t+1} + \cdot\cdot\cdot +(\gamma\lambda)^{T-t-1} \delta_{T-1} \label{eq:At_GAE} \\
    &\text{where}  \; \delta_t = r_t + \gamma V(s_{t+1}) - V(s_t)
\end{align}
With the advantage estimation, the policy gradient loss of the SPPO algorithm can be constructed according to Equation (\ref{scheme1}) and (\ref{scheme2}), respectively.

For Equation (\ref{scheme2}): 
\begin{equation}
L^\pi_t(\theta)=\min \Big( \frac{\pi_{\theta}(a_t|s_t)}{\pi_{\theta^\prime}(a_t|s_t)} \hat{A}^{\pi_{\theta^\prime}}(s_t,a_t), g(\epsilon, \hat{A}^{\pi_{\theta^\prime}}(s_t,a_t)) \Big) + \alpha H(\pi(\cdot|s_t))  \label{eq:Lpi2} 
\end{equation}
where
\begin{align}
 &g(\epsilon, A) = (1+\epsilon)A  \;\text{if} \; A>0  \;\text{else} \; (1-\epsilon)A.
\end{align}

For Equation (\ref{scheme1}): 
\begin{equation}
L^\pi_t(\theta)=\min \Big( \frac{\pi_{\theta}(a_t|s_t)}{\pi_{\theta^\prime}(a_t|s_t)} A^{\pi_{\theta^\prime}}(s_t,a_t), g(\epsilon, A^{\pi_{\theta^\prime}}(s_t,a_t)) \Big) \label{eq:Lpi1} 
\end{equation}
where
\begin{align}
 &  A^{\pi_{\theta^\prime}}(s_t,a_t) = \hat{A}^{\pi_{\theta^\prime}}(s_t,a_t)- \alpha \log  \pi_{\theta^\prime}(a_t|s_t),
\end{align}

We summarize the complete algorithm in Algorithm \ref{algo:1}. 

\begin{algorithm}
\DontPrintSemicolon
\SetAlgoLined
\SetKwInOut{Input}{Input}\SetKwInOut{Output}{Output}
\Input{Initial policy parameters $\theta$ \\
\BlankLine
value-function parameters $\phi$ 
\BlankLine
Temperature $\alpha$ \\
}
 
\BlankLine
\While{not converge}{    
    
    \BlankLine
    
    \For{each step}{
        \BlankLine
        $a_t \sim \pi_{\theta}(a_t|s_t)$ \tcp*{Sample action from the policy}
        \BlankLine
        $s_{t+1} \sim p(s_{t+1}|s_t, a_t)$\tcp*{Sample transition from the environment}
        \BlankLine
        $\mathcal{D} \leftarrow \cup \{(o_t, s_t, a_t, r(s_t, a_t), o_{t+1}, s_{t+1}\}$ \tcp*{collect set of trajectories}
        
        \BlankLine
        Compute rewards-to-go $\hat{R}_t$
        \BlankLine
        Compute advantage estimates, $\hat{A}_t$(using and method of advantage estimate based on the current value function ) $V_{\phi_{k}}$
        
        \BlankLine
        Compute policy gradient $L^\pi(\theta)$ from equation(\ref{eq:Lpi2}) or (\ref{eq:Lpi1})
        \BlankLine
        $\theta \leftarrow \theta - \lambda_{\theta} \bigtriangledown_{\theta} L^\pi(\theta)$ \tcp*{Update policy parameter} 
        \BlankLine
        Compute value gradient from equation(\ref{eq:Lv}) 
        \BlankLine
        $\phi \leftarrow \phi - \lambda_{\phi} \bigtriangledown_{\phi} L^v(\phi)$ \tcp*{Update value parameter} 
    }}
    
\caption{SPPO}
\label{algo:1}
\end{algorithm}

\section{Experiments}\label{header-n4}

Empirically, on-policy algorithms are more suited for parallel and distributed reinforcement learning architecture. Previous off-policy algorithms like SQN and SAC1 are off-policy algorithms. In this paper, we show that the advantages of entropy regularization exploited in SQN and SAC1 can also be applied to on-policy algorithms. In the paper \cite{SAC1}, the authors demonstrate the superiority of SAC1, comparing with DDPG. Now we know that SAC1 and SDDPG are identical. We can infer that reinforcement learning algorithms can benefit from the entropy regularization.

\begin{figure}
    \centering
    \includegraphics[width=0.79\textwidth]{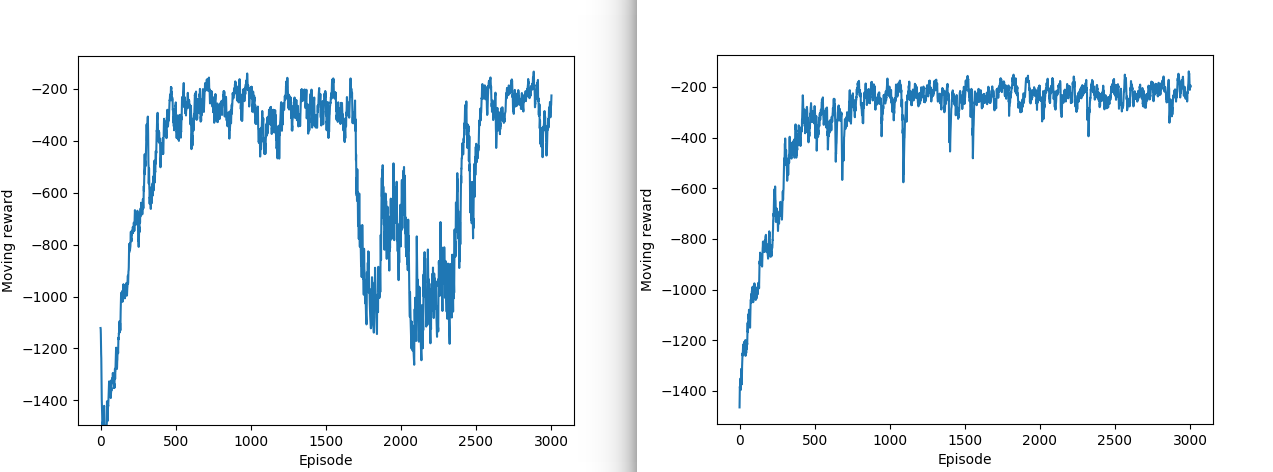}
    \caption{Gym "pedulum-v0" experiments. The left penal shows the moving reward along the training episodes of PPO, while the right one is for SPPO.}
    \label{fig:pendulum}
\end{figure}

We experiment on OpenAI gym environment "pedulum-v0" as a starter, the code is available at \url{https://github.com/createamind/DRL/tree/master/algorithm_standalone/PPO}. This environment is simple yet significant to manifest the advantages of entropy regularization. From Fig.\ref{fig:pendulum}, we can see the superiority of SPPO over PPO, training is more stable and performance is more robust.

We also experiment on gym atari environment "breakout". We implement SPPO by modifying the PPO code of \url{https://github.com/openai/baselines/tree/master/baselines/ppo2}. We find that SPPO can solve "breakout" and get full score of 864 for every episode, while PPO fails. All the hyper-parameters are chosen by defualt values from the original PPO code.
\begin{figure}
    \centering
    \includegraphics[width=0.7\textwidth]{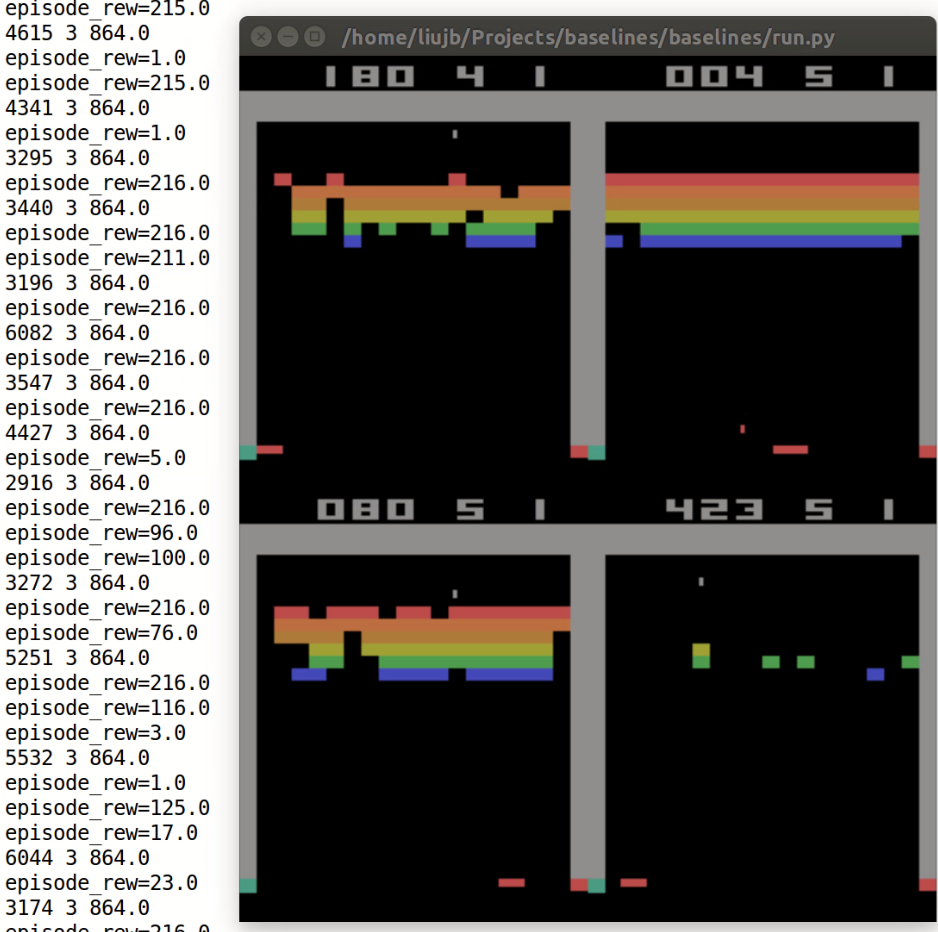}
    \caption{Gym atari environment "breakout" experiments. SPPO can solve "breakout" and get full score of 864, while PPO fails.}
    \label{fig:breakout}
\end{figure}

\subsection{Temperature and Reward Scale}\label{header-n432}
Reward design plays an important role in reinforcement learning. Here we discuss the connection between reward scale and the temperature parameter $\alpha$ in soft policy gradient. The temperature $\alpha$ is actually the reward scale of the entropy bonus. Thus it is the ratio of the temperature $\alpha$ and reward scale that determine the extent of exploration. $\alpha$ can be tuned for a specific environment. We can also adapt $\alpha$ during training by targeting a given entropy for the policy, as suggested in \cite{SAC1}.

\subsection{Local Variance Scale}\label{header-n431}

\begin{wraptable}{r}{0.4 \textwidth}
\begin{longtable}[c]{@{}clll@{}}
\caption{All valid update schemes} 
\label{table:scheme} \\
\toprule
scheme & $\mu$ & $\sigma$\tabularnewline
\midrule
\endhead
1 & \emph{MLP} & \emph{Global $\sigma$} \tabularnewline
2 & \emph{MLP} & \emph{Local $\sigma_s$}\tabularnewline
3 & \emph{MLP} & \emph{Global $\sigma$ * Local $_\Delta\sigma_s$}\tabularnewline
4 & \emph{MLP} & \emph{Global $\sigma$ * Local $\text{clip}(_\Delta\sigma_s)$}\tabularnewline
\bottomrule
\end{longtable}
\end{wraptable} 

In standard policy optimization algorithms like PPO, The policy network uses a global action variance $\sigma$ when action space is continuous. While in SPPO, we try to break this constraint and endow policy networks with more representation capacity. It is a naive way to let the policy network output a local $\sigma$ for every state. This is problematic due to the high update variance of policy gradient. Instead, we maintain the global action variance $\sigma$ as the main source of action variance, but in the meanwhile, we introduce a local variance scale $_\Delta\sigma_s$ for each state. Furthermore, In order to reduce the effect of encouraging the agent to seek an undesired large local variance scale, we propose a new scheme that clips the local variance scale to be logarithmic negative. As shown in Table \ref{table:scheme} we list the four kinds of schemes. We find that it can work collaboratively with the idea of entropy regularization. 

\section{Conclusion}\label{header-n5}

In this work, We  extend the idea of entropy regularization into the on-policy realm and propose the soft policy gradient theorem. With SPGT, We derive the soft version of all the policy optimization algorithms in the standard reinforcement learning. We find that SDDPG and SAC1 are actually identical. Furthermore, the fact reveals the equivalence between soft Q-learning and soft policy gradient. We show that the impact of entropy regularization goes beyond providing the agent with extra exploration. Instead, it also serves a more stable training process by avoiding collapsing. Empirical results show that SPPO outperforms state-of-the-art on-policy PPO algorithm by a substantial margin, providing a promising avenue for reinforcement learning with great robustness and stability.

\bibliographystyle{apalike}
\bibliography{ref.bib}

\end{document}